\documentclass{bvm} 

\begin{document}

\newcommand{\bvmyear}{2026}

\selectlanguage{english} 

\title{Tracking Cancer Through Text}

\subtitle{Longitudinal Extraction From Radiology Reports Using Open-Source Large Language Models}

\titlerunning{Tracking Cancer Through Text}


\author{
\fname{Luc} \lname[0009-0005-6272-2257]{Builtjes} \affiliation{RUMC} \authorsEmail{luc.builtjes@radboudumc.nl} \street{Geert Grooteplein Zuid} \housenumber{10} \zipcode{6525 GA} \city{Nijmegen} \country{The Netherlands} \isResponsibleAuthor,
\fname{Alessa} \lname[0000-0002-7602-803X]{Hering} \affiliation{RUMC} \authorsEmail{alessa.hering@radboudumc.nl} \street{Geert Grooteplein Zuid} \housenumber{10} \zipcode{6525 GA} \city{Nijmegen} \country{The Netherlands}
}

\authorrunning{Builtjes \& Hering}

\institute{
Department of Medical Imaging, Radboudumc, Nijmegen, The Netherlands
}

\email{luc.builtjes@radboudumc.nl}

\maketitle

\begin{abstract}
Radiology reports capture crucial longitudinal information on tumor burden, treatment response, and disease progression, yet their unstructured narrative format complicates automated analysis. While large language models (LLMs) have advanced clinical text processing, most state-of-the-art systems remain proprietary, limiting their applicability in privacy-sensitive healthcare environments. We present a fully open-source, locally deployable pipeline for longitudinal information extraction from radiology reports, implemented using the \texttt{llm\_extractinator} framework. The system applies the \texttt{qwen2.5-72b} model to extract and link target, non-target, and new lesion data across time points in accordance with RECIST criteria. Evaluation on 50 Dutch CT Thorax/Abdomen report pairs yielded high extraction performance, with attribute-level accuracies of 93.7\% for target lesions, 94.9\% for non-target lesions, and 94.0\% for new lesions. The approach demonstrates that open-source LLMs can achieve clinically meaningful performance in multi-timepoint oncology tasks while ensuring data privacy and reproducibility. These results highlight the potential of locally deployable LLMs for scalable extraction of structured longitudinal data from routine clinical text.
\end{abstract}

\section{Introduction}
Radiology reports are a central means of documenting and communicating oncologic findings between specialists, particularly between radiologists and oncologists. They describe tumor burden, treatment response, and disease progression according to standardized criteria such as the Response Evaluation Criteria in Solid Tumors (RECIST) \cite{3871-therasse2000new}. Beyond their role in clinical decision-making, these reports represent a vast, underutilized source of longitudinal information on cancer progression. If automatically extracted, such information could provide large-scale, high-quality data for clinical research, enable retrospective population analyses, and support the development and evaluation of AI models across diverse patient cohorts. Yet, the narrative nature and heterogeneous structure of radiology reports make this information difficult to access in a systematic and scalable way.

Early information extraction efforts relied on rule-based algorithms \cite{3871-draelos2021chest} or supervised deep learning models \cite{3871-tushar2021classification}, but these approaches often lacked generalizability and required substantial data annotation and engineering resources. The emergence of large language models (LLMs) has markedly advanced the ability to process unstructured clinical text without the need for task-specific model training \cite{3871-bigolin2025leavs}. Nevertheless, many state-of-the-art LLMs are proprietary, raising important privacy, security, and reproducibility concerns in healthcare contexts. Open-source LLMs offer a compelling alternative, as they can be deployed locally to ensure that sensitive patient data remain within institutional boundaries.

Current clinical text extraction efforts largely treat radiology reports as independent documents, even though oncologic interpretation is inherently longitudinal. Changes in lesion size and appearance across timepoints determine treatment response under RECIST, and processing reports at the patient level enables linking of corresponding findings across follow-up studies, capturing disease evolution and ensuring consistent interpretation.

\begin{figure}
    \centering
    \includegraphics[width=1.0\linewidth]{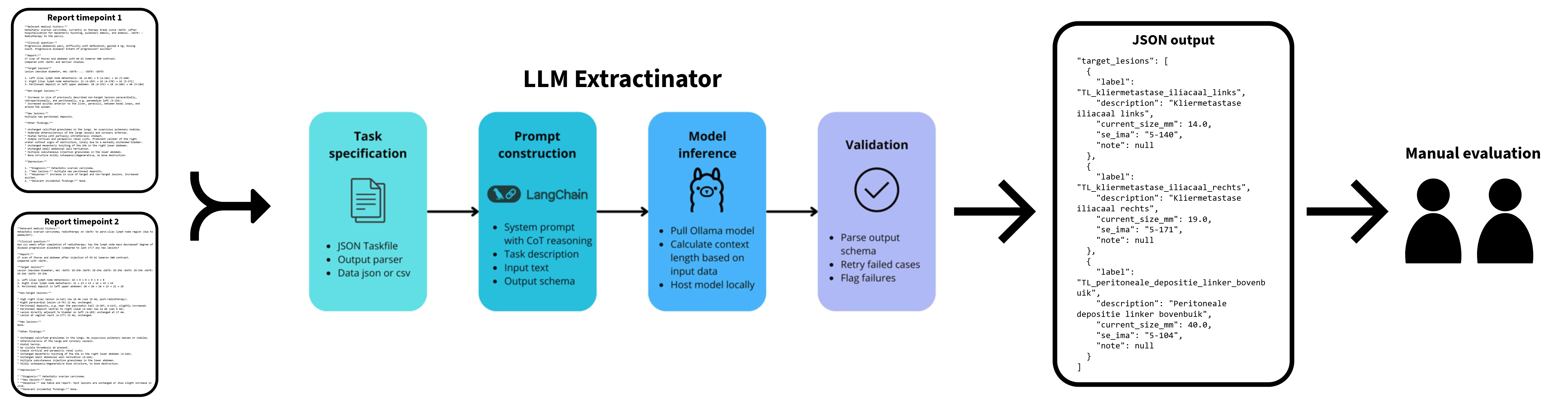}
    \caption{Overview of the study design. Fifty pairs of Dutch thorax/abdomen radiology reports were analyzed using the \texttt{llm\_extractinator} package \cite{3871-builtjes2025llm}, configured with tailored prompts and a customized output parser. The generated structured JSON outputs, representing all lesions extracted from each report pair, were manually assessed for quality by two independent readers.}
    \label{3871-fig:overview}
    \altText{Flowchart illustrating the study design. Fifty paired Dutch thorax/abdomen radiology reports are processed using the llm_extractinator package with customized prompts and an output parser. The system generates structured JSON files containing extracted lesion information for each report pair. These outputs are then independently reviewed and manually assessed for quality by two human readers.}
\end{figure}

In this work, we present a fully open-source and locally deployable pipeline for longitudinal information extraction from Dutch CT Thorax/Abdomen radiology reports. Using the language-agnostic \texttt{llm\_extractinator} framework \cite{3871-builtjes2025llm} with the \texttt{qwen2.5-72b} model, the system extracts and links target lesions (TLs), non-target lesions (NTLs), and new lesions (NLs) across timepoints in accordance with RECIST guidelines. To our knowledge, this is the first demonstration of LLM-based longitudinal extraction from radiology reports. We contribute (i) a reproducible open-source pipeline for clinical text extraction using locally deployable LLMs, (ii) a new task formulation for multi-timepoint lesion linkage under RECIST, and (iii) a quantitative evaluation demonstrating high extraction accuracy (>93\% attribute-level) while preserving data privacy and reproducibility. An overview of our study design is shown in Fig.\ref{3871-fig:overview}.

\section{Materials and methods}

\subsection{Dataset}
As a staring point, we retrieved all radiology reports from patients who received a thoracic and/or abdominal CT scan at the Radboudumc between March 2021 and March 2025. To focus on longitudinal data, we first identified patients with more than one report available across distinct timepoints. Within this subset, we performed a text-based search to identify cases in which the term “target” appeared in at least two reports for the same patient, ensuring the inclusion of longitudinal references to target lesions. Following this automated filtering, 60 report pairs were manually reviewed to confirm their suitability for inclusion. The curated dataset was subsequently partitioned into two subsets: 10 pairs allocated as a debug set for prompt engineering and refinement, and 50 pairs reserved as the final test set for evaluation.

\subsection{LLM\_Extractinator}
We employed the open-source framework \texttt{llm\_extractinator} \cite{3871-builtjes2025llm} to perform structured information extraction. This framework provides a language-agnostic, locally deployable pipeline for clinical text processing using open-source LLMs. It enables users to define extraction tasks through JSON configuration files, specify the model backend, and enforce structured outputs via schema validation (e.g., using Pydantic models).

After identifying the relevant report pairs for each patient, we processed these pairs using the Extractinator pipeline to produce structured outputs suitable for downstream quantitative analyses. We selected \texttt{qwen2.5:72b-instruct-q4\_K\_M} as the underlying large language model (LLM) following a comparative evaluation on a debug dataset, where it achieved the best overall performance among models compatible with our computational constraints. All inferences were conducted on two NVIDIA A100 GPUs (40 GB each) with the temperature parameter set to 0

\subsubsection{Prompt engineering}

The final prompt was designed to ensure accurate and longitudinally consistent extraction of lesion information. Reports were always processed in concatenated pairs, enabling the model to recognize recurring lesions and maintain temporal continuity. For the LLM, this meant reasoning across both reports to determine whether each lesion persisted, resolved, or appeared newly, while still extracting only the \emph{current timepoint} measurements. 

The prompt included an in-context example illustrating the Dutch RECIST table format and its corresponding structured output. This improved both schema adherence and temporal label stability.

Key prompt rules reflecting RECIST conventions were summarized as follows: (i) in tabular data, the \emph{rightmost} numeric value immediately preceding the last valid series--image (\texttt{SE-IMA}) reference represents the current measurement; (ii) valid \texttt{SE-IMA} identifiers match the pattern \texttt{\textasciicircum\textbackslash d\{1,3\}-\textbackslash d\{1,4\}\$} and are never interpreted as sizes; (iii) lesions are categorized as \emph{target}, \emph{non-target}, or \emph{new}, with prose-only findings included using \texttt{null} sizes when appropriate; (iv) each lesion receives a stable label (\texttt{TL\_}, \texttt{NTL\_}, or \texttt{NL\_}) derived from its anatomical description to ensure consistent tracking; and (v) outputs must use integer millimetre sizes.

\subsubsection{Output parser design}
The output parser serves as a blueprint guiding the LLM to produce JSON outputs that strictly adhere to a predefined structure, as specified by a Pydantic~v2 schema. It consists of three nested models: \emph{Lesion} containing \texttt{label}, \texttt{description}, \texttt{current\_size\_mm}, \texttt{se\_ima}, and \texttt{note}, \emph{Report} grouping TLs, NTLs, and NLs under a unique StudyInstanceUID, and \emph{OutputParser} aggregating multiple reports. All fields are optional so the model can opt to return an empty list if no lesions in a given category are present.

\subsection{Experiments}

To evaluate the LLM’s performance in longitudinal lesion extraction, its outputs were compared with manual annotations from two independent readers across 50 patient report pairs. Both readers independently identified all TLs, NTLs, and NLs in each report. The goal was to assess how accurately the LLM detected, linked, and characterized corresponding lesions over time.

Each lesion was evaluated on three attributes per report: (i) \emph{label consistency}: correct lesion identification and naming across timepoints; (ii) \emph{size accuracy}: correct extraction of lesion measurements; and (iii) \emph{series/image (se\_ima) accuracy}: correct retrieval of DICOM identifiers for lesion localization.

For TLs and NLs, each attribute received a binary label (“correct”/“incorrect”), yielding six total attributes per lesion pair. For NTLs, 20 cases were evaluated at the attribute level and 30 at the report level for efficiency. Reader annotations were pooled to compute overall accuracy, and inter-reader variability was quantified by lesion-level agreement on identification and correspondence.

\begin{figure}
    \centering
    \includegraphics[width=1.0\linewidth]{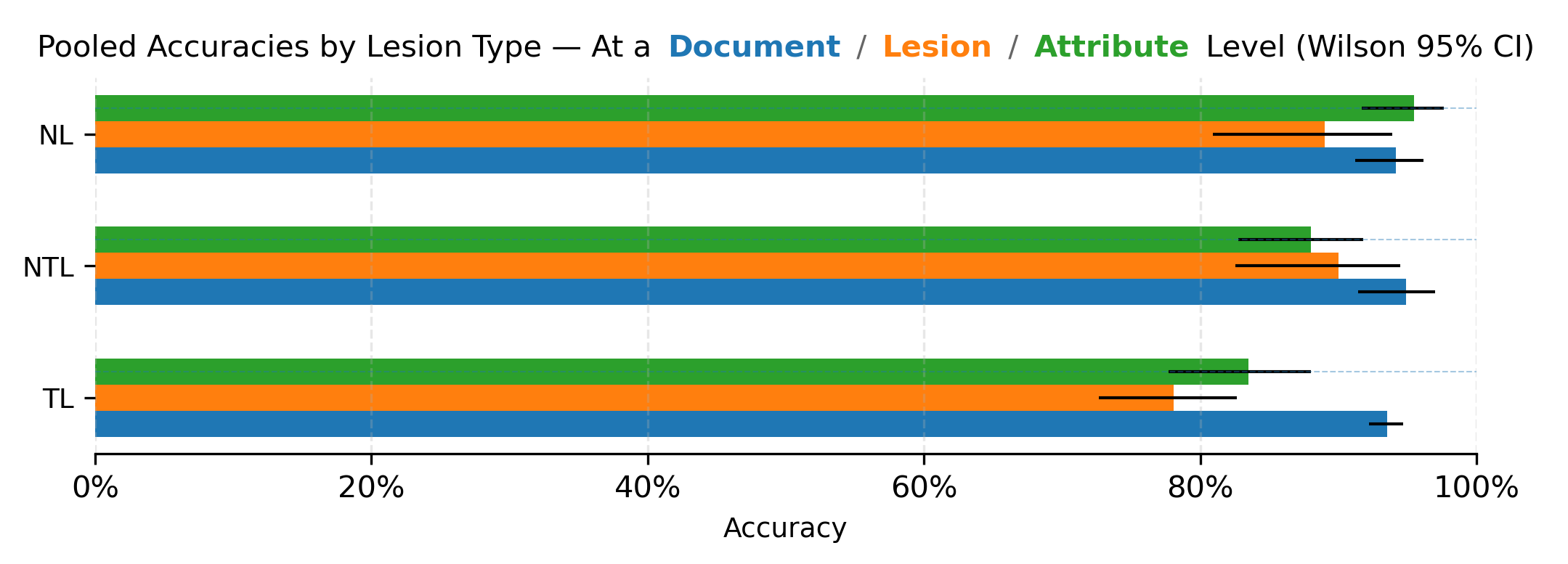}
    \caption{Comparison of LLM extraction accuracy across lesion types: Target Lesions (TL), Non-Target Lesions (NTL), and New Lesions (NL). Accuracy is shown at three levels: document (green), lesion (orange), and attribute (blue).}
    \label{3871-fig:results}
    \altText{Grouped bar chart comparing LLM extraction accuracy for three lesion categories: Target Lesions, Non-Target Lesions, and New Lesions. For each category, three bars represent accuracy at the document level (green), lesion level (orange), and attribute level (blue), showing differences in performance across evaluation levels.}
\end{figure}

\section{Results}

A total of 50 longitudinal report pairs (100 individual reports) were evaluated. On average, each report pair contained 2.6 TLs, 5.0 NTLs, and 0.91 NLs. A visual overview of the results is shown in Fig.~\ref{3871-fig:results}.

Across both readers, the pooled accuracy for complete TL extraction was 78.1\% (95\% CI: 72.5--83.0). At the attribute level, 93.7\% (95\% CI: 92.2--95.0) of TL attributes were correctly extracted, and 83.5\% (95\% CI: 77.4--88.2) of reports contained no TL extraction errors. 

For NTLs, performance was similarly high, with a pooled accuracy of 90.0\% (95\% CI: 84.9--93.7) for correct identification. The attribute-level accuracy reached 94.9\% (95\% CI: 92.5--96.6), and 88.0\% (95\% CI: 82.1--92.0) of reports showed no NTL extraction errors. 

Extraction of NLs demonstrated comparable performance, with 89.1\% (95\% CI: 80.7--94.2) of lesions identified without error. Attribute-level accuracy was 94.0\% (95\% CI: 91.1--96.1), and 95.5\% (95\% CI: 91.2--97.8) of reports were free from NL extraction errors.

Each longitudinal report pair contained on average 30 total attributes; in 62.0\% of pairs (95\% CI: 52.2–70.9), all attributes were extracted without any errors.

Inter-reader agreement was high overall at a \emph{93.3\%} agreement lesion-level agreement rate. Differences between readers were small and not statistically significant for any category (two-proportion $z$-test, $p > 0.05$).

\section{Discussion}

This study demonstrates that open-source large language models can achieve high performance in extracting and linking longitudinal lesion data from radiology reports. Using the \texttt{llm\_extractinator} framework with the \texttt{qwen2.5-72b} model, we obtained attribute-level accuracies above 93\% across lesion types. These results highlight that open and locally deployable LLMs can reach a level of reliability suitable for clinical research while ensuring data privacy and reproducibility.

The longitudinal setting introduces challenges distinct from single-report extraction, as the model must not only identify findings but also link them coherently across timepoints. Labeling consistency between timepoints was excellent for target lesions (TLs), where all label attributes were correct. This indicates that the stable label design worked effectively for longitudinal tracking. For non-target lesions (NTLs), consistency was somewhat lower, which likely reflects the more variable wording used by radiologists across follow-up reports rather than model limitations. For example, lymph nodes could be described collectively (e.g., “multiple lymph nodes”) in one report and individually in another, leading to apparent inconsistencies in label matching despite correct underlying detection.
Additionally, the handling of the “Other findings” section was occasionally inconsistent, with the model sometimes classifying them as NTLs and other times omitting them.

The distribution of target lesion errors was balanced, with 45\% involving the size attribute and 55\% the \texttt{se\_ima} identifier. This even split is expected, as errors in one attribute often coincide with errors in the other when interpreting tabular data, highlighting that the model’s understanding of lesion identity was generally robust. While the model showed resilience to differences in report layout, it occasionally struggled when tabular sections were wrapped across lines or split into multiple parts, sometimes selecting the last value from the first segment instead of continuing correctly. These formatting-related errors are understandable given the loss of visual table cues in plain-text form, and providing more specific in-context examples could further improve consistency in such cases.

Lesions described as difficult to measure posed a specific challenge. They could be indicated in various ways, such as by using an asterisk with a corresponding footnote under the table or by simply noting ``nm'' (not measurable) without further context. The model sometimes failed to return a \texttt{null} value when no measurement was provided, or conversely returned \texttt{null} despite an approximate value being available. Nevertheless, it often handled these edge cases correctly, showing that the model can interpret nuanced clinical phrasing with reasonable consistency. Similar variability was observed when lesions were indicated as resolved with a dash (``--''), suggesting that additional examples or explicit rules could further improve stability in these scenarios.

Overall, the presented pipeline performs strongly in a demanding multi-timepoint clinical text task using fully open-source components. Despite minor challenges related to formatting, measurement ambiguity, and heterogeneous report structure, the approach achieves accurate, reproducible, and privacy-preserving longitudinal information extraction. These findings support the feasibility of open LLMs for clinical natural language processing and highlight their potential to enable scalable data extraction in oncology.

\section{Acknowledgements}
This publication is part of the project OncoChange with file number 21121 of the research program Veni - Applied and Engineering Sciences. Which is (partly) financed by the Dutch Research Council (NWO)

\printbibliography

\end{document}